\documentclass[runningheads]{llncs}
\usepackage[T1]{fontenc}
\usepackage{siunitx}
\usepackage{fancyhdr}

\usepackage{graphicx}
\usepackage{siunitx}
\usepackage{subcaption}
\usepackage{adjustbox}
\begin{document}
\title{Advancing Precision in Multi-Point Cloud Fusion Environments}
%\title{Evaluation of 3D Point Cloud Distances: A Comparative Study in Multi-Point Cloud Fusion Environments}
%j
\author{Ulugbek Alibekov\inst{1}\and
Vanessa Staderini\inst{1} \and
Philipp Schneider\inst{1} \and
Doris Antensteiner\inst{1}}
\authorrunning{U. Alibekov et al.}
\institute{AIT Austrian Institute of Technology, Vienna, Austria }
\maketitle              % typeset the header of the contribution
\begin{abstract}
%THIS SHORT ABSTRACT IS A PLACEHOLDER INCLUDING SOME IDEAS; PLEASE EXTEND AND WRITE A FULL ABSTRACT FOR THIS WORK
 
This research focuses on visual industrial inspection by evaluating point clouds and multi-point cloud matching methods. We also introduce a synthetic dataset for quantitative evaluation of registration method and various distance metrics for point cloud comparison. Additionally, we present a novel CloudCompare plugin for merging multiple point clouds and visualizing surface defects, enhancing the accuracy and efficiency of automated inspection systems. 

%KEY WORDS TO HELP WRITING THE INTRODUCTION: Quality Control, Industrial Inspection, 3D Reconstruction, Point Cloud Matching, Multi-Point Cloud Registration, Noise Reduction, Surface Defects Detection, Iterative Closest Point (ICP), Pose Graph, CloudCompare Plugin, Vision Sensors, Synthetic Dataset, Distance Metrics, High-Precision Registration, Automated Inspection Systems

\end{abstract}
\section{Introduction} \label{sec:introduction}% ADD AN % ADD AN OVERVIEW PICTURE AT THE START AND REFERENCE TO IT IN THE INTRODUCTION
With the rapid development of the manufacturing industry, the need for robust quality control inspection is increasing significantly. Currently, most quality inspection procedures are performed manually by humans. This manual process is often slow, not always reliable, and requires substantial financial investment due to labor costs. 

\begin{figure}
  \centering
  \includegraphics[width=0.6\textwidth]{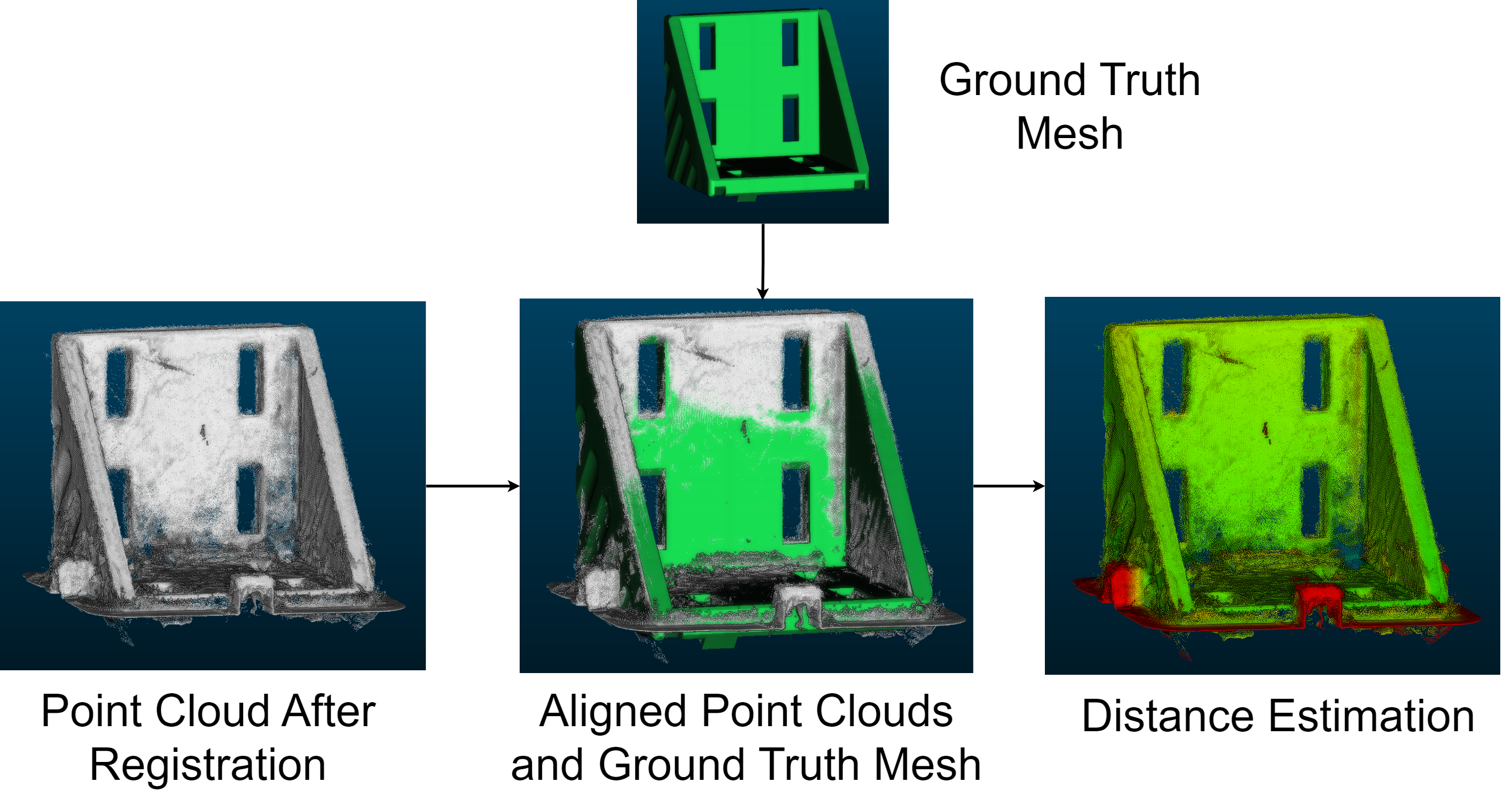} % Adjust the width as needed
  \caption{We begin the inspection of the metal bracket by acquiring partial scans of the object. These scans are then processed using the Refined Pose Graph method to ensure accurate alignment. Next, we align the processed scans with the ground truth mesh of the bracket. Finally, we perform distance estimation relative to the ground truth model to identify any potential defects. }
  \label{fig:overview}
\end{figure}

Automated industrial pipelines using customized algorithms offer a more efficient alternative. They are faster, more consistent, and require less investment in the long run \cite{mpcrnet}. As a result, many companies are looking to incorporate advanced vision technologies, including robotic-assisted inspection, into their quality control systems to ensure higher standards of their manufacturing processes~\cite{huo2023research}.

%INSERT GOAL HERE. E.g.: 
In this work, we extend our research in \cite{alibekov2024evaluation} to focus on improved registration for which we created a novel CloudCompare plugin. Our objective was to evaluate various point-cloud distance metrics and analyze the "Pose Graph" and "Global ICP" registration methods using both synthetic and real data. Building on that foundation, this current work extends our research by focusing on enhancing the Pose Graph method. Additionally, we transformed our implementation into a CloudCompare plugin and tested it across different scenarios. 

For a machine to understand and analyze a 3D object during inspection, the object should be scanned into a special data structure called a point cloud. A point cloud is a set of points in 3D space, with each point representing a specific part of the surface of the object \cite{wu2022inenet}. Various vision methods to obtain point clouds include  structured light sensor, laser scanning, stereo matching, and photogrammetry, each producing a partial point cloud that may include color information about the object~\cite{lee2000novel}.

When acquiring partial point clouds, several challenges arise. Noise is introduced during the acquisition process, which can detrimentally affect the image data. Also, there are often missing parts of the object that were not correctly captured in the point cloud. Sources for these errors include challenging reflection properties of the object's surface and sensor issues. Moreover, the partial point clouds are usually not aligned in space due to sensor errors and offsets \cite{huang2023cross}. Therefore, reliable 3D registration methods are needed to merge multiple point clouds and address these issues.

\textbf{Key contributions of this work}: 
\begin{enumerate}
    \item \textbf{Refined Pose Graph Method:} We introduce an improved Pose Graph method that significantly improves the accuracy of registering multiple partial point clouds.
    \item \textbf{Parameter Investigation:} We conducted an investigation into the selection and interconnections of registration parameters, providing valuable insights for optimizing the registration process.
    \item \textbf{CloudCompare Plugin Implementation:} We developed and implemented a CloudCompare plugin that is specifically designed for multiview point cloud registration.
\end{enumerate}

The paper is organized as follows: In Section \ref{sec:related_works}, we present the state-of-the-art methodologies related to point cloud acquisitions and registration. We then present our methodology and the corresponding results in Sections \ref{sec:methodology} and \ref{sec:results}. Finally, the conclusions and an outlook are provided in Section \ref{sec:conclusions}.

\section{Related Works}\label{sec:related_works}

%How point clouds can be acquired
Point cloud acquisition encompasses various vision-based technologies, including laser-based and camera-based approaches. Laser-based systems, such as LiDAR, emit laser beams towards the object and measure the time it takes for the light to return, determining the 3D location of each laser hit. This information is then transformed into a point cloud \cite{wandinger2005introduction}. Camera-based methods involve using multiple images to reconstruct a 3D representation of an object. Techniques like stereo vision and photogrammetry analyze differences between images taken from different angles to generate point clouds. Structured light sensors project a light pattern onto the object, which deforms on the basis of the object's geometry. By analyzing these deformations, the sensor can obtain a 3D representation of the object \cite{thorstensen2021high}. The experimentally acquired point clouds are often misaligned due to sensor errors and drift of the acquisition system. For this reason it is necessary to accurately align the point clouds by applying a registration method.

\subsection{Classical Registration Methods}
Most classical registration approaches are based on the Iterative Closest Point (ICP) algorithm \cite{924423}. It consists of five main stages: (i) Selection - identifying points in the overlapping regions between pairs of point clouds; (ii) Correspondence - finding a corresponding subset of points in the other point cloud using the selected points; (iii) Rejection - eliminating false or incorrect corresponding pairs; (iv) Weighting - assigning a weight to each corresponding pair of points; and (v) Optimization - performing an optimization algorithm to find the rigid transformation matrix that minimizes the weighted and squared distances between corresponding points. This procedure is repeated iteratively until a desired error threshold is achieved. For our purpose, two variants of the ICP algorithm were chosen: Global ICP \cite{glira2015a} and Pose Graph \cite{choi2015robust}.

The Global ICP method introduces various strategies for selecting points in one point cloud (using random or uniform sampling), identifying corresponding pairs of points, weighting these pairs, removing outliers, and computing the rigid transformation that minimizes the distance between points. Unlike traditional methods that register pairs of point clouds, this approach focuses on aligning multiple point clouds simultaneously. Moreover, using voxel downsampling, Global ICP can effectively handle large point cloud data.

The Pose Graph method is a variant of the ICP algorithm that represents multiple point clouds as a pose graph. In this framework, the nodes are partial scans and the edges connect two nodes. If the nodes are neighboring, meaning the point clouds are close to each other, the edge connecting them is called an odometry edge. For non-neighboring nodes, a loop closure edge is used. Each edge contains a transformation matrix to align the source point cloud to the target. The method registers pairs of local scene fragments, constructing a global model based on these alignments, while removing low-confidence pairs based on point cloud density. By performing an optimization on the entire pose graph, it is possible to minimize distances between all partial point clouds and obtain an aligned 3D representation.

\subsection{Deep Learning Methods}
According to a survey on learning-based point cloud registration methods, these methods can be divided into correspondence-free and correspondence-based categories \cite{MONJIAZAD202358}. The survey concluded that correspondence-free methods often struggle with global feature differences in point clouds, while correspondence-based methods fail when faced with missing correspondences. Deep learning methods excel at finding coarse initial transformations between point clouds. They are particularly effective at identifying distinct point feature representations, which is advantageous when dealing with symmetric or repetitive elements, low overlap ratios, or weak geometric features (e.g., flat objects) \cite{sarode2019pcrnet}. However, these methods are primarily focused on scene reconstruction tasks in which submillimetric precision is not required. For this reason, classical approaches offering better spatial resolutions are more suitable for quality inspection tasks \cite{brightman2023point}.

\subsection{Point Cloud Distance Measurements} 

%comparing point clouds to other point clouds and measuring the distance 

Once the merged point cloud is obtained, it has to be compared to the ground truth to identify any imperfections and defects. Depending on the nature of the ground truth model, there are two primary distance metrics available: cloud-to-cloud and cloud-to-mesh. For the cloud-to-cloud distance, both the reference and the source are point clouds, whereas for the cloud-to-mesh distance, one is a point cloud and the other one is a mesh.

Cloud-to-cloud distance metrics can be categorized into point-to-point or point-to-plane methods. 

For point-to-point, the distance is computed between each point in one point cloud and its corresponding point in the other. In our study, we chose to analyze the most popular point-to-point distance metrics, including Chamfer Distance \cite{wu2021density}, Hausdorff Distance \cite{huttenlocher1993comparing}, and Earth Mover's Distance \cite{yuan2018pcn}. Chamfer Distance computes the average sum of squared distances between corresponding pairs of points, Hausdorff Distance finds the maximum distance between any pair of nearest neighbor points, and Earth Mover's Distance establishes a one-to-one correlation between points in the point clouds to minimize the distance between corresponding points. While Earth Mover's Distance ideally provides an accurate distance estimation, it requires significant computational power and point clouds of the same size. In reality, the application of this metric is often impossible due to presence of noise and outliers.

For point-to-plane distance metrics, the distance is measured from a point in one point cloud to the fitting plane passing through a set of nearest neighbors. For our evaluation, we chose the Least Squares \cite{peterson2009k}, Quadratic \cite{chernov2011least}, and Triangulation methods \cite{chen2004optimal}. As the name suggests, the Least Squares method computes the best-fitting plane through the nearest neighbors using the least squares approach. It then finds the distance from a point of interest in one point cloud to the fitting plane by projecting a vector from the point to the plane and computing the normal vector. The Quadratic method works similarly to the Least Squares method, but the best-fitting plane is calculated using a quadratic equation. In the case of Triangulation, a set of triangles is created by connecting k-nearest neighbors in 2D using Delaunay triangulation. This results in a 2.5D mesh, where the distance is computed by finding the closest triangle to the point and performing a point-to-plane distance estimation.

Cloud-to-mesh distance metrics compute the distance from a point cloud to a reference mesh, where a mesh consists of vertices, edges, and faces that represent the shape of a 3D object. Typically, the faces of the mesh are created using triangles, quadrilaterals, or other simple polygons \cite{cobb2009mesh}. The distance estimation is similar to point-to-plane distance metrics with the plane being represented by a mesh. If the orthogonal projection of the point onto the plane falls outside the triangle, the distance to the nearest edge is used. Depending on the direction of the normal vector, the distance can be positive (i.e., point is outside the mesh) or negative (i.e., point is inside the mesh) \cite{jones19953d}.

%HIGHLIGHT ALSO THE NEW ICI CC PLUGIN AND WHICH GAP IT IS FILLING - CURRENTLY SUCH A PLUGIN DOESNT EXIST AND IS HIGHLY USEFUL FOR INDUSTRIAL INSPECTION AND DEFECT DETECTION

\section{Methodology}\label{sec:methodology}

\subsection{Synthetic Data Generation}
\begin{figure}[h!]
  \centering
  \begin{subfigure}[b]{0.47\textwidth}
    \centering
    \includegraphics[height=0.5\textwidth]{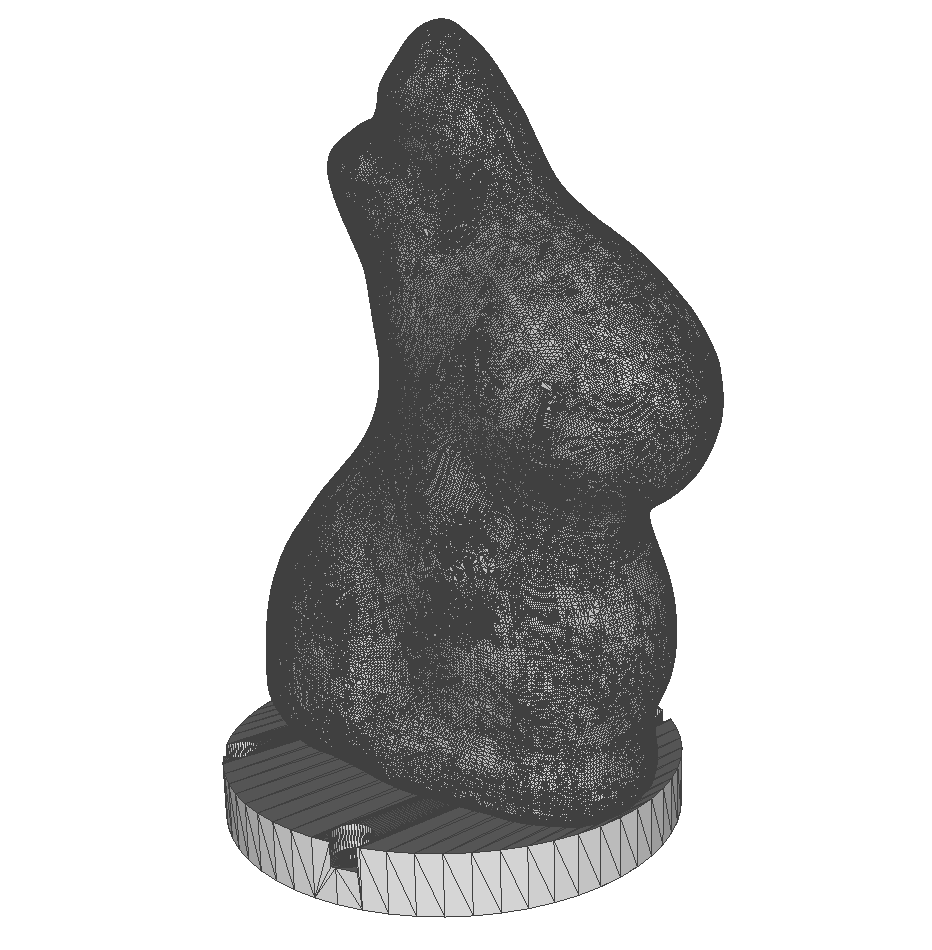}
    \caption{3D mesh of a bunny used for synthetic dataset generation of partial scans. This model was later used to 3D-print the object and use it during real-world experiments.}
    \label{fig:meshrabbit}
  \end{subfigure}
  \hfill
  \begin{subfigure}[b]{0.47\textwidth}
    \centering
    \includegraphics[height=0.5\textwidth]{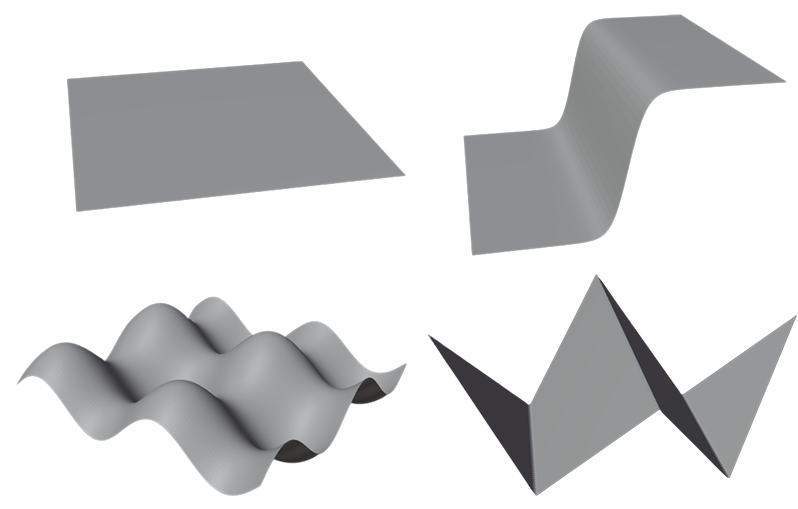}
    \caption{Different synthetically generated meshes used in this work (plane, slope, sine wave, triangular wave). Each mesh represents a different degree of shape complexity.}
    \label{fig:meshes}
  \end{subfigure}
  \caption{Illustrations of the 3D mesh of a bunny used for the generation of partial scans and various synthetically generated meshes from \cite{alibekov2024evaluation}.}
\end{figure}

To objectively compare registration methods, a synthetic dataset was created. A 3D mesh of a bunny was used to simulate a camera taking scans from different viewpoints (see Figure \ref{fig:meshrabbit}). To achieve full coverage of the object, a Poisson disc sampling method was employed to determine optimal camera poses, ensuring the entire surface of the object was captured with a minimal number of viewpoints \cite{staderini2023surface}.

For simulated image acquisition, the partial point clouds were synthetically generated via ray tracing. To align with the real parameters of our lab camera, the sensor model was set to an array of 1920 × 1200 pixels,  a field of view (FOV) spanning $\SI{38.70}{\degree} \times \SI{24.75}{\degree}$, and a depth of field (DOF) between $\SI{350}{ \mm}$ and $\SI{700}{ \mm}$. The synthetically generated partial scans were then rigidly transformed using random translation and rotation values ranging from 0-15 mm and 0-15 degrees, respectively.

Since the ground truth transformations are known, they can be compared with the output of the registration method to calculate the error and evaluate the accuracy of the registration methods.

For point cloud metrics, we synthetically generated four well-defined shapes to compare methods as shown in Figure \ref{fig:meshes}. Specifically, we created meshes of a plane, slope, sinusoidal wave, and a triangular wave using Blender. Each shape was segmented into 16 parts to create a fine mesh, with a top-view size of 1x1 meters.

\begin{figure}
    \centering

    \begin{subfigure}{0.32\textwidth}
        \centering
        \includegraphics[width=\linewidth]{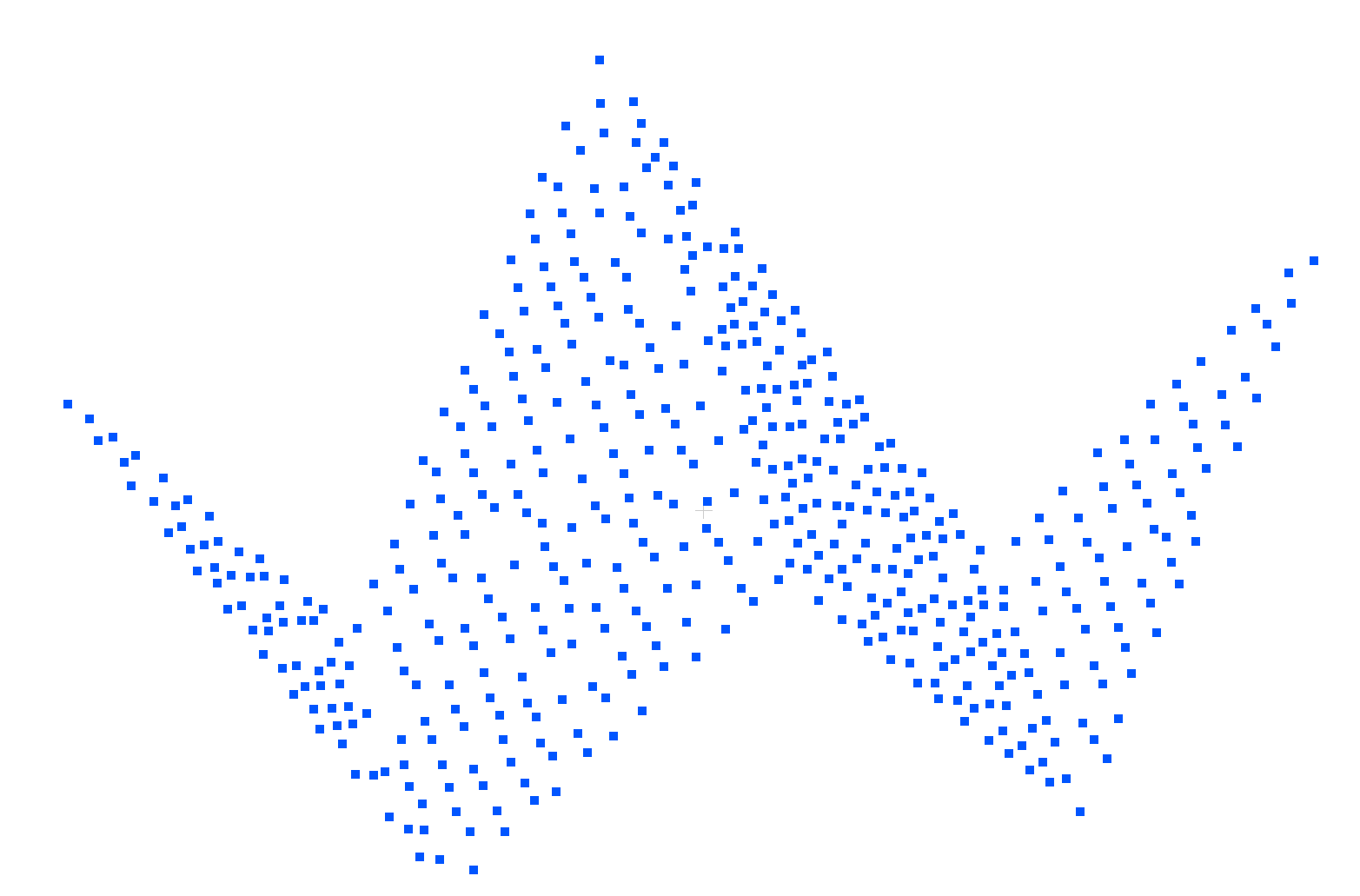}
        \caption{}
        \label{fig:sub1}
    \end{subfigure}
    \hfill
    \begin{subfigure}{0.32\textwidth}
        \centering
        \includegraphics[width=\linewidth]{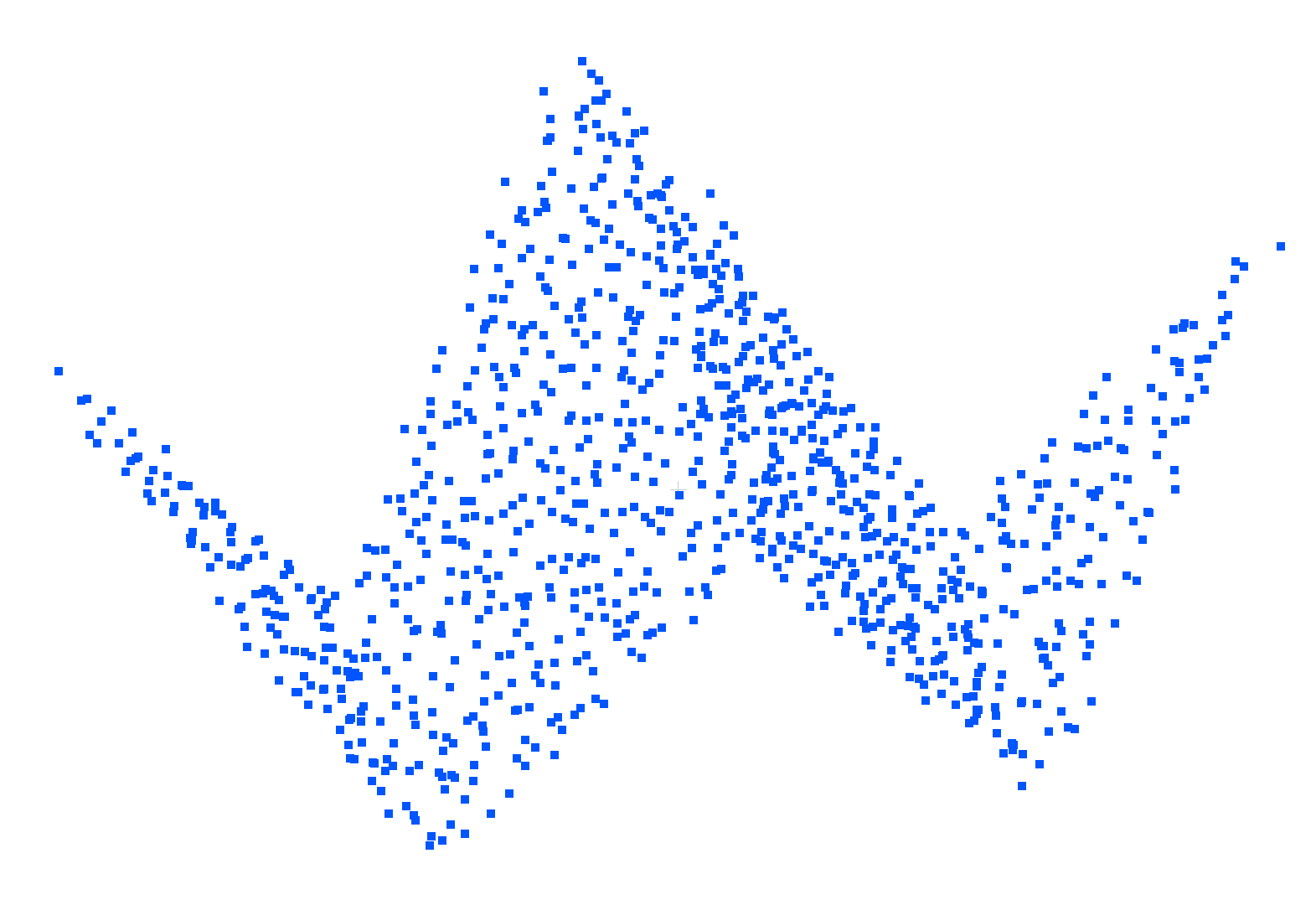}
        \caption{}
        \label{fig:sub2}
    \end{subfigure}
    \hfill
    \begin{subfigure}{0.32\textwidth}
        \centering
        \includegraphics[width=\linewidth]{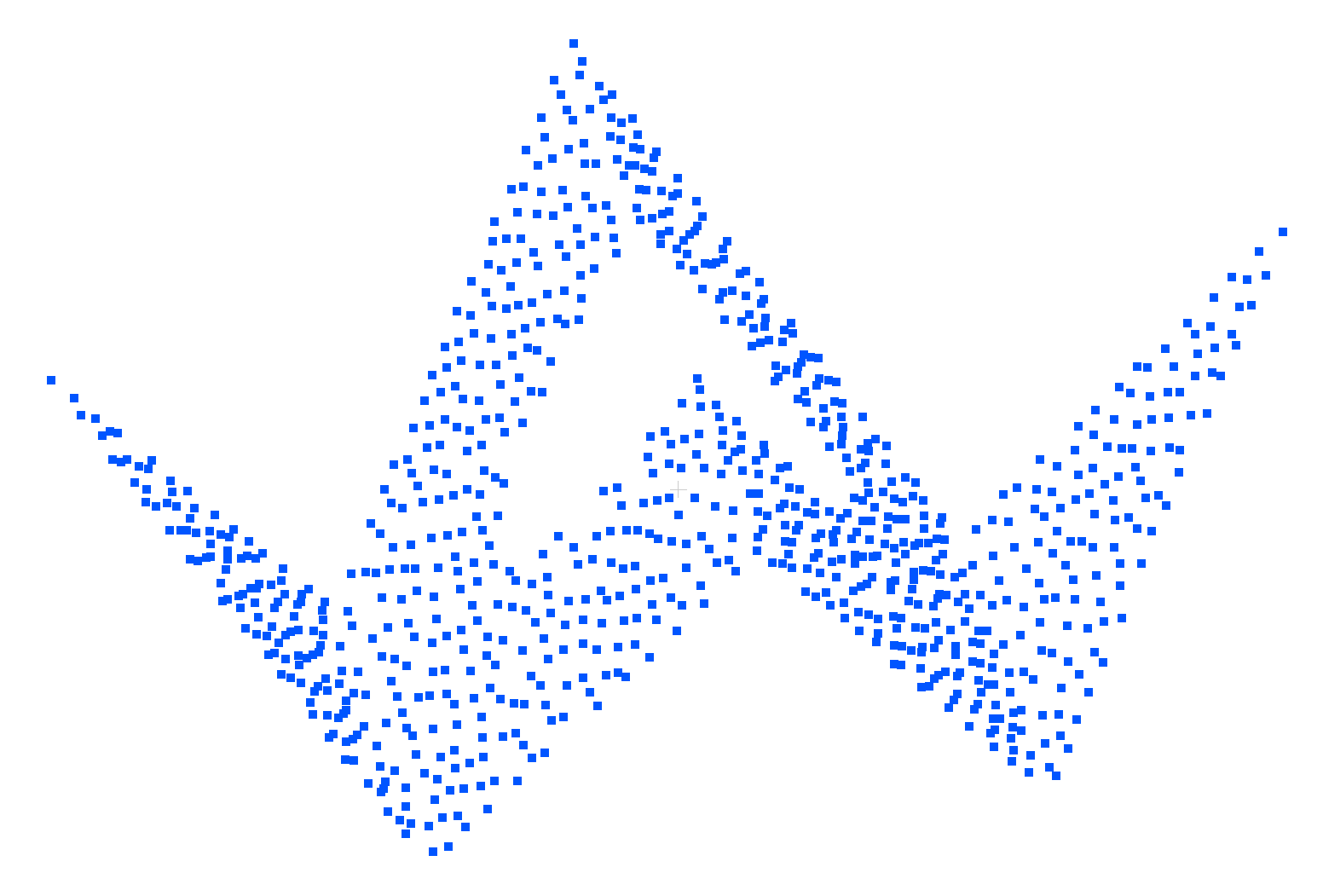}
        \caption{}
        \label{fig:sub3}
    \end{subfigure}
\vspace*{5pt}
    \caption{Triangular waves with various point cloud densities and perturbations from \cite{alibekov2024evaluation}: (a) point cloud density level (adjusted by the sampling factor) of 0.5m, (b) noise with a standard deviation of 0.05m, and (c) hole with a radius of 0.5m.}
    \label{fig:different_tri_shapes}
\end{figure}

Each of the four synthesized shapes was initially sampled with 1000 points. Then, each shape was duplicated and the copy was shifted along the \textit{z}-axis by 0.5 meters, establishing our ground truth distance values. Depending on the scenario (see below), the copied point cloud was influenced by perturbations such as noise and holes, and also by the sampling factor (see Figure \ref{fig:different_tri_shapes}).

For noise, we applied a normal distribution with a mean of 0 and a standard deviation ranging from 0.01 to 0.1 meters. For hole size, points were removed from the centroid location with a variable radius of 0.1 to 1 meters. The (linear) sampling factor indicated the density of the point cloud, where 0 represented the original point cloud and 1 indicated no points.

\subsection{Refined Pose Graph Registration Algorithm}
In the Pose Graph approach, each node is represented by a geometry $P_i$, that is, a point cloud, and it is associated with a transformation matrix $T_i$ to transform $P_i$ to the global frame. This is defined with respect to the first node $P_0$. Thus, $T_0$ is the identity matrix. The objective is to determine the set of remaining unknowns $T_i$. This can be done by collecting transformations between neighboring nodes: The edge connecting the two neighboring nodes $P_i$ and $P_j$ is associated with the respective transformation $T_{i,j}$. In this way, a pairwise registration is conducted.

Previously, the pairwise registration was done using point-to-plane ICP inside Pose Graph method~\cite{alibekov2024evaluation}. This had the disadvantage of not estimating the normal vectors of the point clouds. Due to this, traditional point-to-plane ICP may incorrectly align the points to the two different sides of a wall. In contrast, information on the normal of each point allows for aligning only those points with similar normal directions. With that, the two sides of a wall remain separated and are correctly represented for instance. Therefore, we developed a Generalized ICP method that accounts for such information on surface normals. The Generalized ICP is a variant of the ICP method by Segal et. al. \cite{segal2009generalized}. It attaches a probabilistic model to the minimization step of the ICP. This way, it is possible to incorporate surface information from both scans, by computing the surface covariance matrices. For that, the normal estimation for every point is needed. As a result, the Generalized ICP determines the plane-to-plane distance instead of the point-to-plane distance, which reduces the error caused by incorrectly registered point pairs.

% \subsection{Synthetic Data Generation}
% %While our final goal is to find defects, we first need to create full point clouds out of multiple views / acquisitions. To evaluate the point cloud registration methods we created and identified distance measurements and created a synthetic dataset in order to evaluate the point cloud registration quantitatively ... 
% % 

% \subsubsection{Synthetic Data Generation of Well-defined Shapes}
% %We generated point clouds with well-defined shapes (see Fig.\ref{} for testing several registration algorithms. These synthtetic datasets allow us to add noise and other disturbances systematically for reaching a thorough evaluation.

% \subsubsection{Synthetic Data Generation of Partial Scans}
% %Also created a synthetic dataset for partial scans by simulating real-world conditions. Our introduced defects include misalignments and noise. 

\subsection{Pose Graph Parameters}
In the Pose Graph method, three parameters must be manually set by the user: (i) the voxel size, (ii) the maximum distance threshold, and (iii) the edge prune threshold.

The voxel size controls the size of the voxels during downsampling. A voxel is a cube with a specific size, and combined together they create a voxel grid used to uniformly downsample the point cloud. All points are bucketed into voxels, and a single point is left by averaging all the contained points. By increasing this parameter, more points can be removed from the point cloud, significantly reducing the computational complexity while preserving the point clouds' structure. 

The maximum distance threshold is used to discard the correspondence between points that are far from each other. In  fact, an outlier of a point cloud might be incorrectly paired with a point from another point cloud. This is avoided ensuring that pairs with a distance larger than the maximum distance threshold are excluded. The determination of this parameter is critical as big values would determine many wrong correspondences. On the other hand, a value too low could potentially cause the algorithm to get stuck in a local minimum. 

The edge prune threshold is used to evaluate the edges of the pose graph and remove the so-called outlier edges, i.e., those edges having a transformation matrix that is significantly different from the rest. These edges may be due to different factors such as noise, errors in pairwise registration or incorrect corresponding pairs, and they must be removed before performing global optimization.

%\subsection{CloudCompare Plugin Development}

\subsection{CloudCompare Plugin Development}

\begin{figure}
  \centering
  \includegraphics[width=0.7\textwidth]{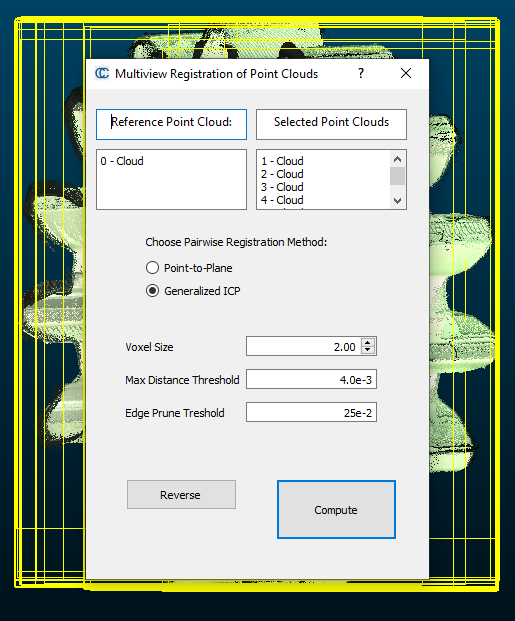} % Adjust the width as needed
  \caption{The Graphical User Interface (GUI) of our multiview registration plugin. The user can swap the source and the reference point clouds, adjust parameters, and select the pairwise registration method. }
  \label{fig:cloudcompareplugin}
\end{figure}

After obtaining a merged point cloud, a visualization and analysis tool is necessary. For this purpose, we used the open-source software CloudCompare \cite{girardeau2016cloudcompare} that offers a wide variety of tools for manually editing and rendering 3D point clouds. CloudCompare includes its own "Fast Global Registration" \cite{rusu20113d} method for aligning multiple partial point clouds. However, when tested with our dataset, this method failed to achieve proper alignment. As a result, we decided to integrated our Refined Pose Graph method as a plugin for CloudCompare (see Figure \ref{fig:cloudcompareplugin}). The graphical user interface of the plugin was created using Qt Designer and the implementation of the method is based on the C++ Open3D library. Our plugin offers two windows: one for displaying the selected point clouds and another for the source point cloud, against which the other point clouds are aligned. The user can choose to perform the registration using the "Point-To-Plane" or the "Generalized ICP" method. A parameter selection section is available to set the appropriate voxel size, maximum distance threshold, and edge prune threshold. Also, "Reverse" button is added, in case if the alignment was wrong and user wants to redo the results. Finally, a "Compute" button can be pressed to start the multiview registration process.

\section{Results and Discussion}\label{sec:results}

\subsection{Comparison of Different Registration methods}
We compared the Global ICP, Pose Graph and Refined Pose Graph methods using our synthetically generated data acquired from different viewpoints in a simulated environment. The comparison was based on the mean value between the ground truth and the estimated transformation matrices.

\begin{table}%[h]
    \centering
    \caption{Average absolute mean errors for Global ICP and Pose Graph registration methods across eight partial point cloud scans.}
    \adjustbox{max width=\columnwidth}{
        \begin{tabular}{|c|c|c|c|c|}
            \hline
            Rotation (degrees) & Translation (mm) & Global ICP & Pose Graph & Refined Pose Graph \\
            \hline
            [0,1] & [0,1] & 0.0045 & 0.0352 & \textbf{0.0028} \\
            \hline
            [1,3] & [1,3] & 0.0416 & 0.0567 & \textbf{0.0047} \\
            \hline
            [3,6] & [3,6] & 0.1668 & 0.1718 & \textbf{0.0286} \\
            \hline
            [6,10] & [6,10] & 0.5482 & 0.5117 & \textbf{0.0456} \\
            \hline
            [10,15] & [10,15] & 0.5280 & 0.5789 & \textbf{0.0780} \\
            \hline
        \end{tabular}
    }
    \label{tab:icpmethods} 
\end{table}

By looking at the numerical results presented in Table \ref{tab:icpmethods}, the Refined Pose Graph significantly outperforms both Global ICP and Pose Graph registration methods. However, as the magnitude of the applied transformations increases, their performance deteriorates. As it was expected, Pose Graph methods are sensitive to parameter settings such as voxel size, maximum distance threshold, and edge prune threshold. Therefore, a parameter investigation step was needed to obtain the accurate results.

\subsection{Parameter Investigation}

We found that for a large point cloud containing 100,000 - 500,000 points, a voxel size of 2 is a good tradeoff between speed and accuracy. However, for point clouds with fewer points it is better to set the voxel size to 1 and skip the downsampling step, as the latter can lead to the loss of points preserving the structure. We identified a dependency between the maximum distance threshold and the voxel size. Once downsampling is complete, the distance between points increases compared to the voxel size, and therefore, it is important to adjust the maximum distance threshold accordingly. Consequently, the maximum distance threshold should be set to $voxel\_size * m$, where $m$ can be within the range of $[1-4]$. Based on our findings, the edge prune threshold is also dependent on the voxel size and should be set as $voxel\_size/p$ with $p \in [2-4]$.

% The following figure illustrates these results, with the X and Y axes representing the average rotation and translation applied to the partial point clouds. The size of the icon in the figure indicates the error, where smaller icons represent smaller errors, and larger icons represent larger errors.

% \begin{figure}[h!]
%   \centering
%   \begin{subfigure}[b]{\textwidth}
%     \centering
%     \includegraphics[height=0.2\textheight]{images/matlab_icp_result.png}
%     \caption{Global ICP.}
%     \label{fig:icp_result}
%   \end{subfigure}
%   \vspace{0.5cm}
%   \begin{subfigure}[b]{\textwidth}
%     \centering
%     \includegraphics[height=0.2\textheight]{images/posegraph1_result.pdf}
%     \caption{Pose Graph.}
%     \label{fig:posegraph_result}
%   \end{subfigure}
%   \vspace{0.5cm}
%   \begin{subfigure}[b]{\textwidth}
%     \centering
%     \includegraphics[height=0.2\textheight]{images/pose_graph_updated.png}
%     \caption{Refined Pose Graph.}
%     \label{fig:refined_pose_graph}
%   \end{subfigure}
%   \caption{Evaluation of Global ICP, Pose Graph, and Refined Pose Graph methods by applying random transformations and rotations to synthetically generated partial scans.}
% \end{figure}

\subsection{Comparison of Cloud-to-Cloud Distance Metrics}

\begin{figure}
  \centering
  \includegraphics[height=0.80\textwidth]{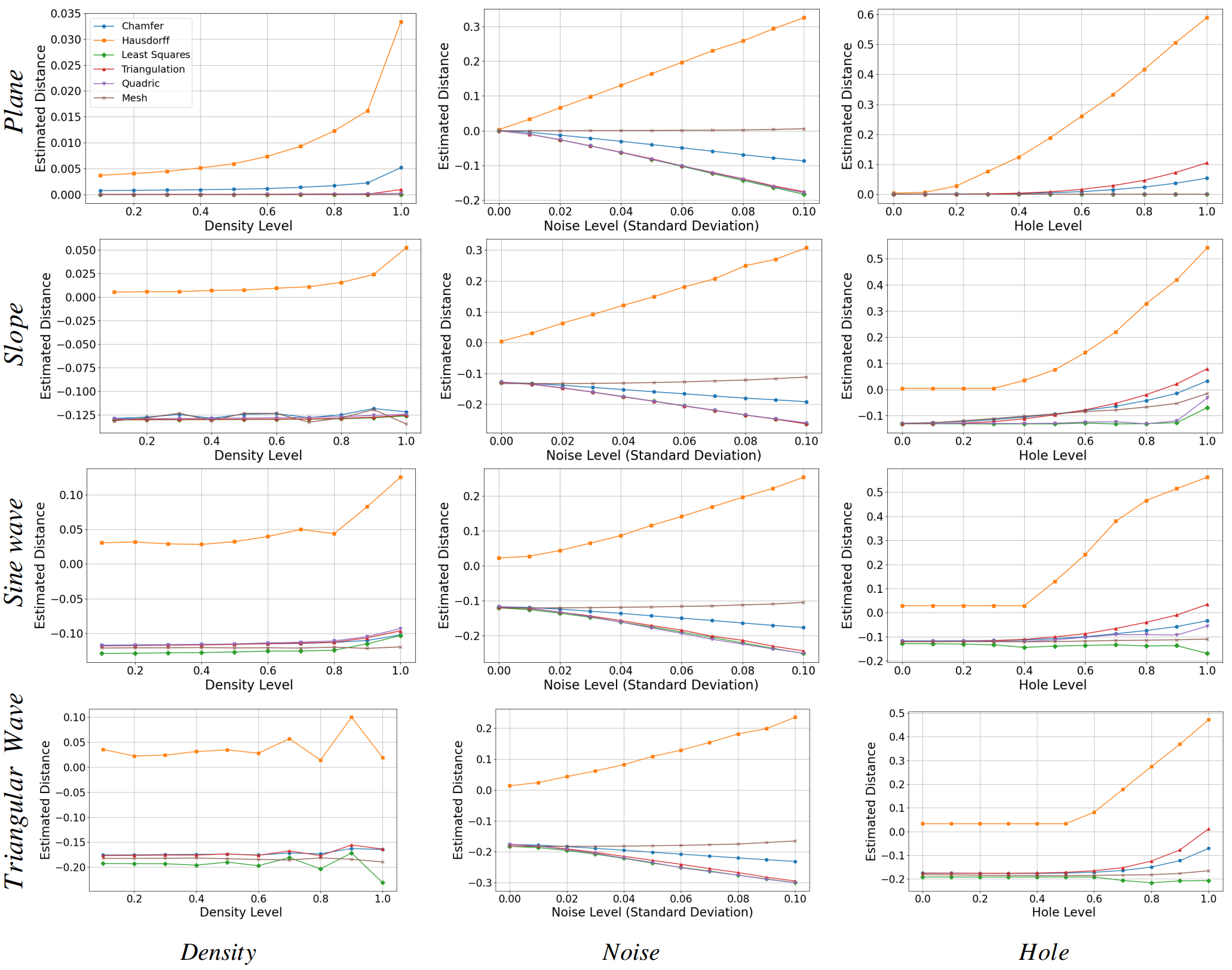} % Adjust the width as needed
  \caption{Evaluation of different point cloud distance metrics in relation to varying noise levels, hole sizes, and for varying point cloud densities from \cite{alibekov2024evaluation}. For each scenario, the results were averaged over 100 executions. The estimated distance is calculated as the deviation from the ground truth value.}
  \label{fig:distance_metrics}
\end{figure}

For point cloud metrics, we computed the distances on synthetically generated data with various perturbations, i.e. different levels and sizes of noise and holes, respectively, and for different point cloud densities (sampling factors), as shown in Fig. \ref{fig:distance_metrics}.  The \textit{y}-axis represents the deviation from the ground truth value, while the \textit{x}-axis indicates the applied perturbations and point cloud densities.

Our evaluation revealed that the shape of the object significantly influences the accuracy of the estimated distances. This is because all methods rely on nearest neighbor searches, and as shape complexity increases, identifying the nearest neighbors becomes more challenging. Additionally, we found that the Hausdorff distance measure is negatively affected by high levels of noise and large holes and sampling factor (i.e., low point cloud densities). The Chamfer Distance showed a similar behavior for point-to-plane distances and was less sensitive to noise due to its averaging procedure. It is important to mention that changing point cloud densitiesand hole sizes did not significantly affect distance estimation. Among all distance metrics, the cloud-to-mesh method demonstrated the most accurate and consistent results across different noise levels, hole sizes, and sampling factors.

\subsection{CloudCompare Plugin Tests}
After developing the GUI and implementing the Refined Pose Graph algorithm, we compiled and tested our new plugin within CloudCompare. The plugin performs multiview registration, where users can directly adjust parameters such as the voxel size, the maximum distance threshold, and the edge prune threshold within the plugin interface.

\subsection{Real-world Data Acquisition for Defect Detection}
%Introduction: We want to find defects on objects, hence we will have a reference point cloud or CAD object model. The goal is to compare the acquired point cloud of the object with our reference. The acquired point cloud can consist of multiple point clouds could result of the previous (multi-)point cloud registration process described in Sec.~ref{}, especially in the case of robotic multi-view inspection. For this comparison we use a linear inspection setup instead (INSERT IMAGE OR DRAWING OF STANDARD ICI AND REFERENCE TO IT). To achieve a real world acquisition, we use a linear stage (NAME OF STAGE), a camera (NAME OF CAMERA) and create a point cloud using multi-view and photometric stereo information as described in https://www.mdpi.com/1424-8220/18/2/431. 

\begin{figure}
  \centering
  \includegraphics[width=0.5\textwidth]{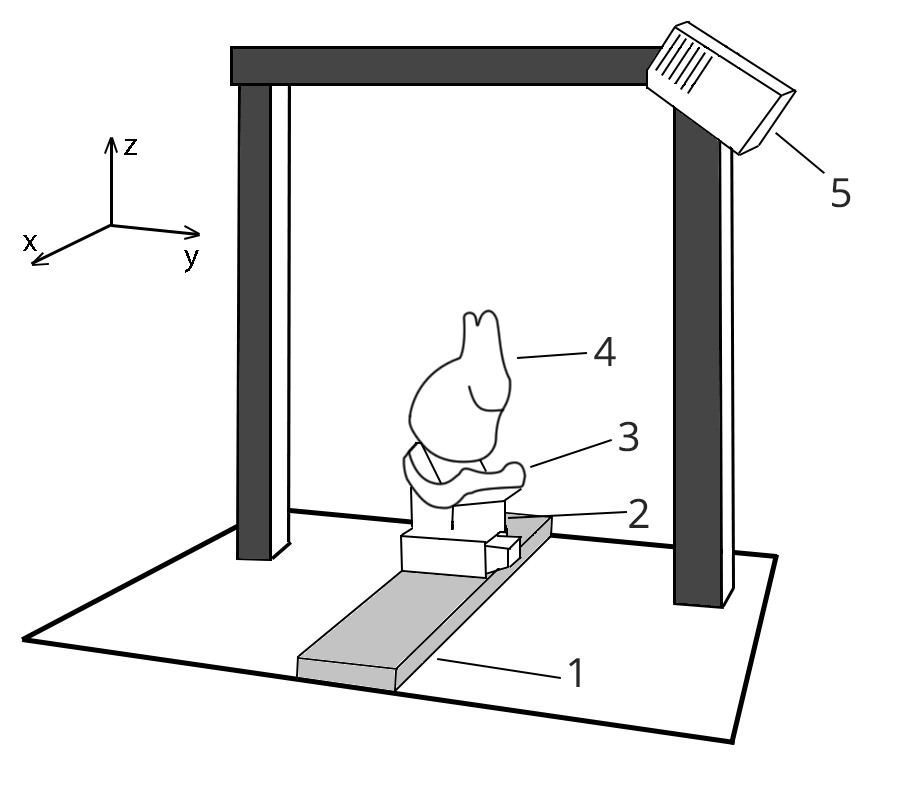} % Adjust the width as needed
  \caption{Schematics of the gantry lab setup for inspection from \cite{alibekov2024evaluation}. Here, 1 - moving linear stage, 2 - rotating stage, 3 - tilt stage (goniometer), 4 - moving object, 5 - structured light sensor. }
  \label{fig:setup}
\end{figure}

%Initial Alignment
When acquiring point clouds, an initial alignment is essential  to register the point clouds. This can be achieved by either extracting features from markers placed in the inspection scene or by utilizing the available kinematic information from the gantry or robotic system that moves the object and/or sensor during the inspection. We utilized the forward kinematics information from our gantry lab setup for inspection  (see Figure \ref{fig:setup}), which consists of a linear stage, a rotation stage, and a tilt stage (goniometer). In this configuration, the structured light sensor is fixed while the inspected object is moved by the setup. By calibrating the sensor and applying the forward kinematics of the setup, we were able to achieve a good initial alignment of the partial point clouds.

The partial scans were obtained using a Zivid One+ S structured light sensor. To obtain full coverage of the model, a method proposed by  Staderini et al. \cite{staderini2023surface} was used. Optimal viewpoint generation was employed to get the sensor-object poses. 

\begin{figure}
    \centering

    \begin{subfigure}{0.32\textwidth}
        \centering
        \includegraphics[width=\linewidth]{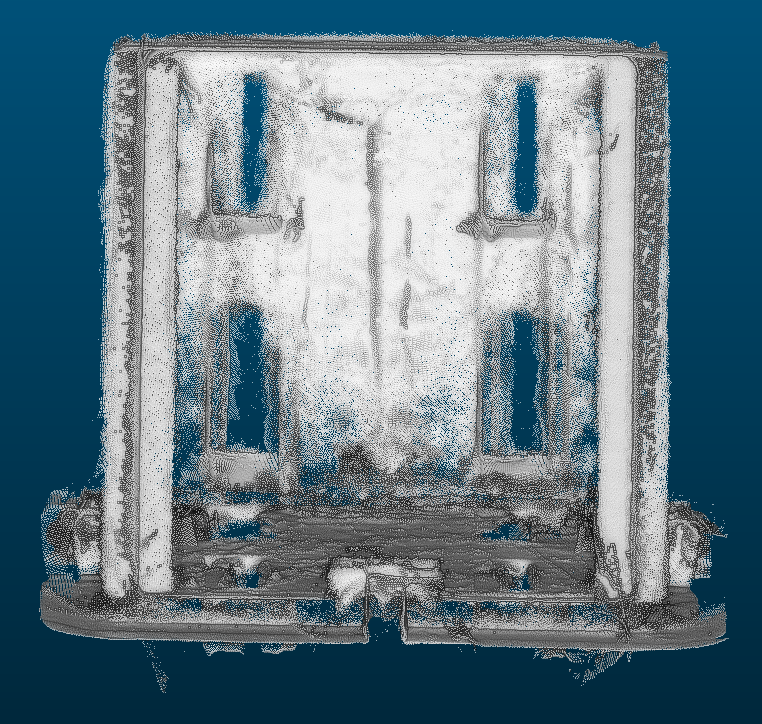}
        \caption{}
        \label{fig:sub1}
    \end{subfigure}
    \hfill
    \begin{subfigure}{0.32\textwidth}
        \centering
        \includegraphics[width=\linewidth]{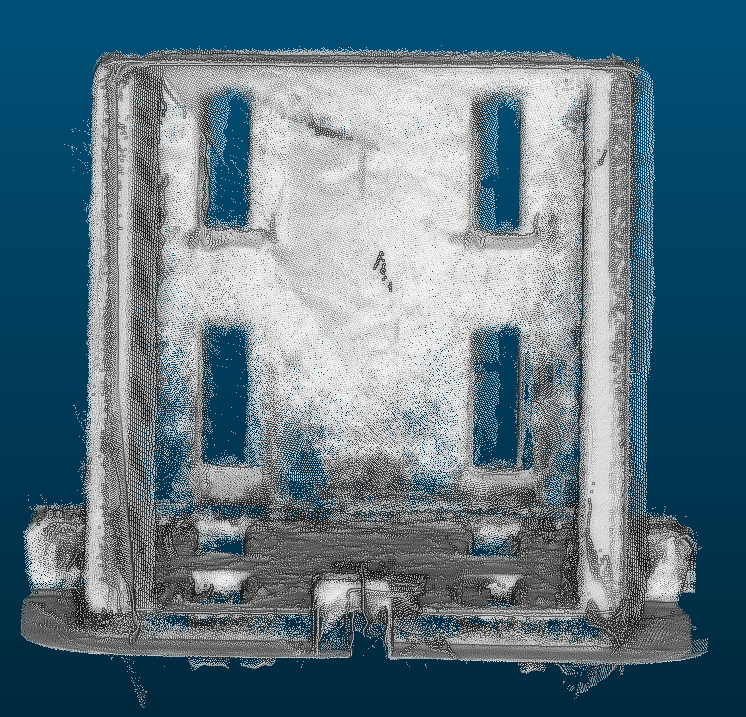}
        \caption{}
        \label{fig:sub2}
    \end{subfigure}
    \hfill
    \begin{subfigure}{0.32\textwidth}
        \centering
        \includegraphics[width=\linewidth]{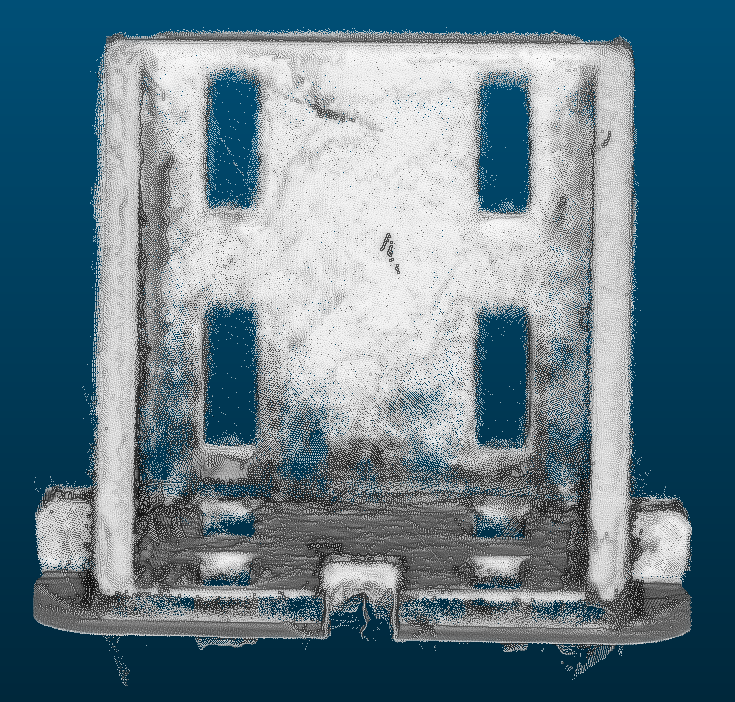}
        \caption{}
        \label{fig:sub3}
    \end{subfigure}
\vspace*{5pt}
    \caption{Aligned point clouds using different registration methods: (a) Global ICP, (b) Pose Graph, and (c) Refined Pose Graph.}
    \label{fig:registration}
\end{figure}

After acquiring the partial scans of the metal bracket object, a preprocessing step was necessary to remove the background stage and noise. This involved cropping the point clouds based on the bounding box and applying a statistical outlier filter to eliminate outliers. Once preprocessing was complete, the registration step was performed, as illustrated in Figure \ref{fig:registration}.

As seen in the images, the Global ICP and Pose Graph methods did not align the point clouds accurately, resulting in significant misalignment between the object's walls. This issue arises because the standard point-to-plane algorithm doesn't account for the direction of normals, leading to incorrect alignment of opposite sides of the wall. In contrast, the Refined Pose Graph method considers the direction of normals, allowing for proper alignment of the walls, as demonstrated in the image (c) of Figure \ref{fig:registration}. 

\begin{figure}
    \centering

    \begin{subfigure}{0.32\textwidth}
        \centering
        \includegraphics[width=\linewidth]{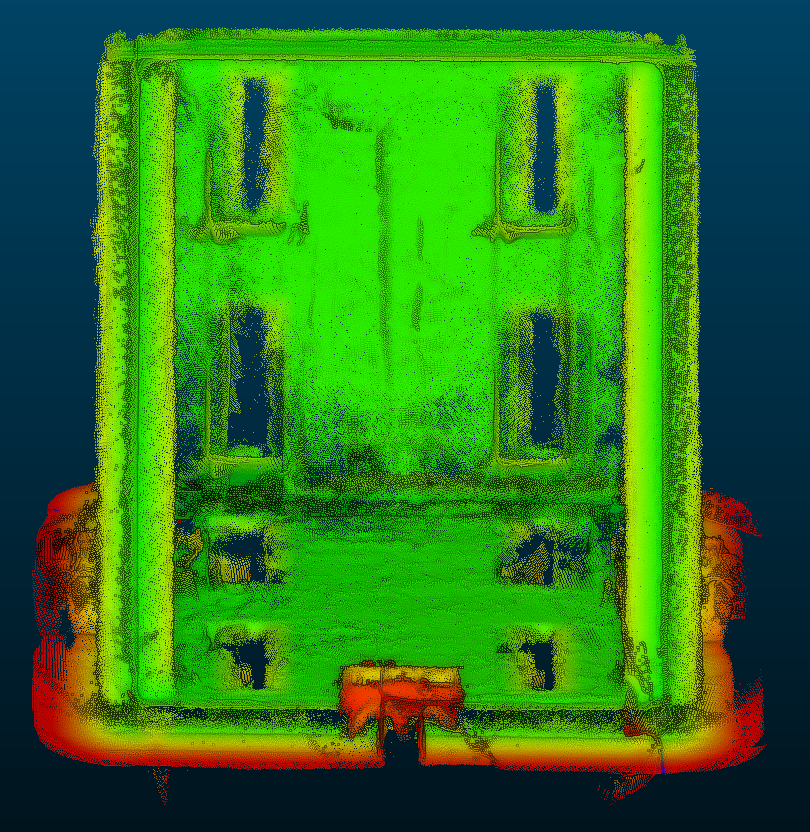}
        \caption{}
        \label{fig:sub1}
    \end{subfigure}
    \hfill
    \begin{subfigure}{0.32\textwidth}
        \centering
        \includegraphics[width=\linewidth]{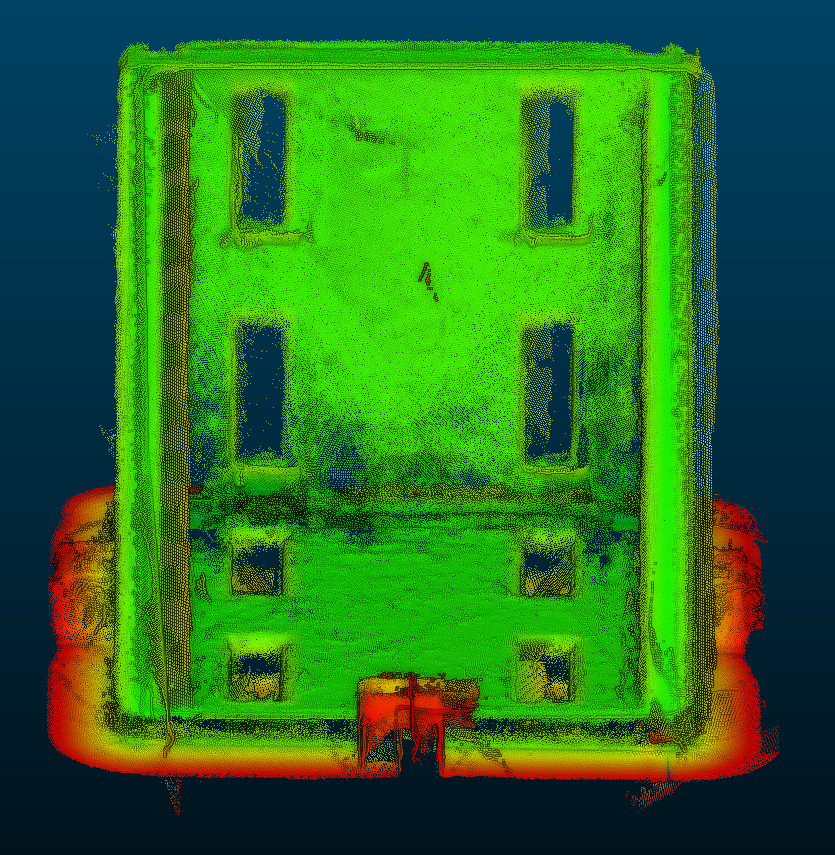}
        \caption{}
        \label{fig:sub2}
    \end{subfigure}
    \hfill
    \begin{subfigure}{0.32\textwidth}
        \centering
        \includegraphics[width=\linewidth]{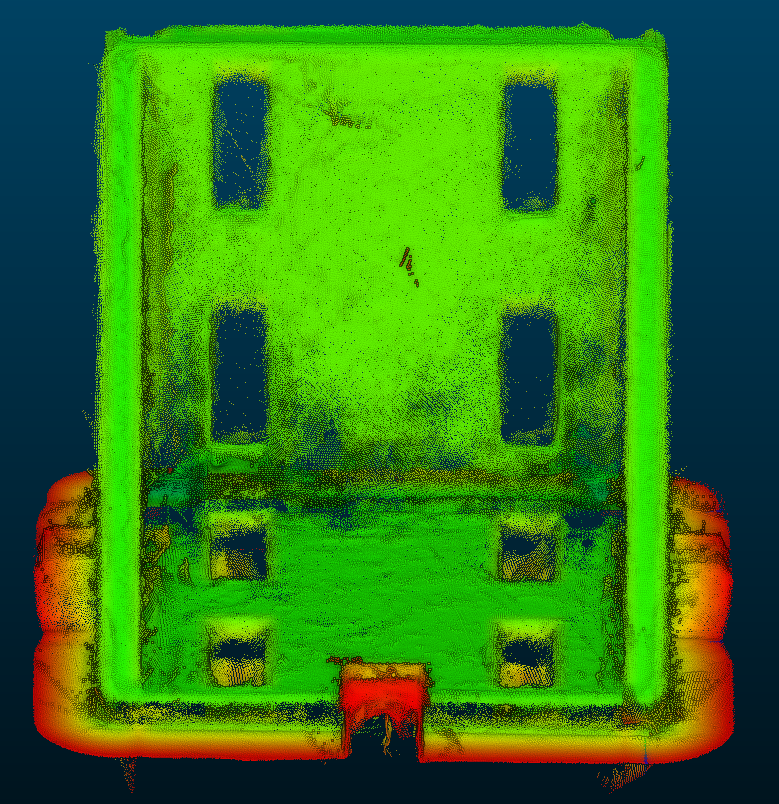}
        \caption{}
        \label{fig:sub3}
    \end{subfigure}
\vspace*{5pt}
    \caption{Distance map obtained by applying Cloud-To-Mesh distance estimation after performing registration using (a) Global ICP, (b) Pose Graph, and (c) Refined Pose Graph.}
    \label{fig:distance}
\end{figure}

Each merged point cloud was compared to the ground truth mesh, starting with an alignment process. Following alignment, a Cloud-to-Mesh Distance Estimation method was applied to calculate the distances between the point cloud and the mesh (see Figure \ref{fig:distance}). In the visualization, red indicates a large positive deviation from the ground truth, while blue represents a large negative deviation. Green indicates distances close to zero.

\begin{figure}
    \centering

    \begin{subfigure}{0.32\textwidth}
        \centering
        \includegraphics[width=\linewidth]{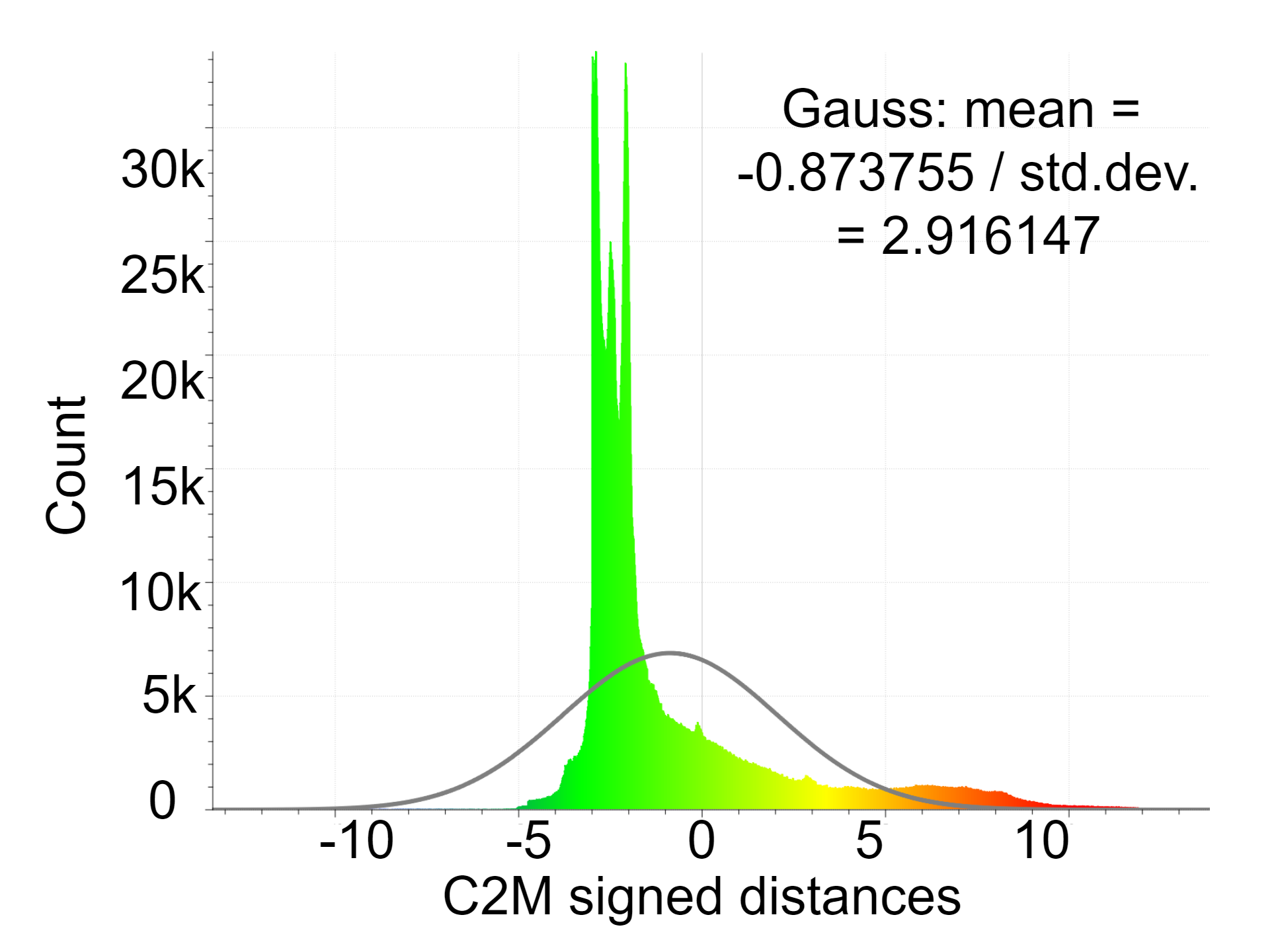}
        \caption{}
        \label{fig:sub1}
    \end{subfigure}
    \hfill
    \begin{subfigure}{0.32\textwidth}
        \centering
        \includegraphics[width=\linewidth]{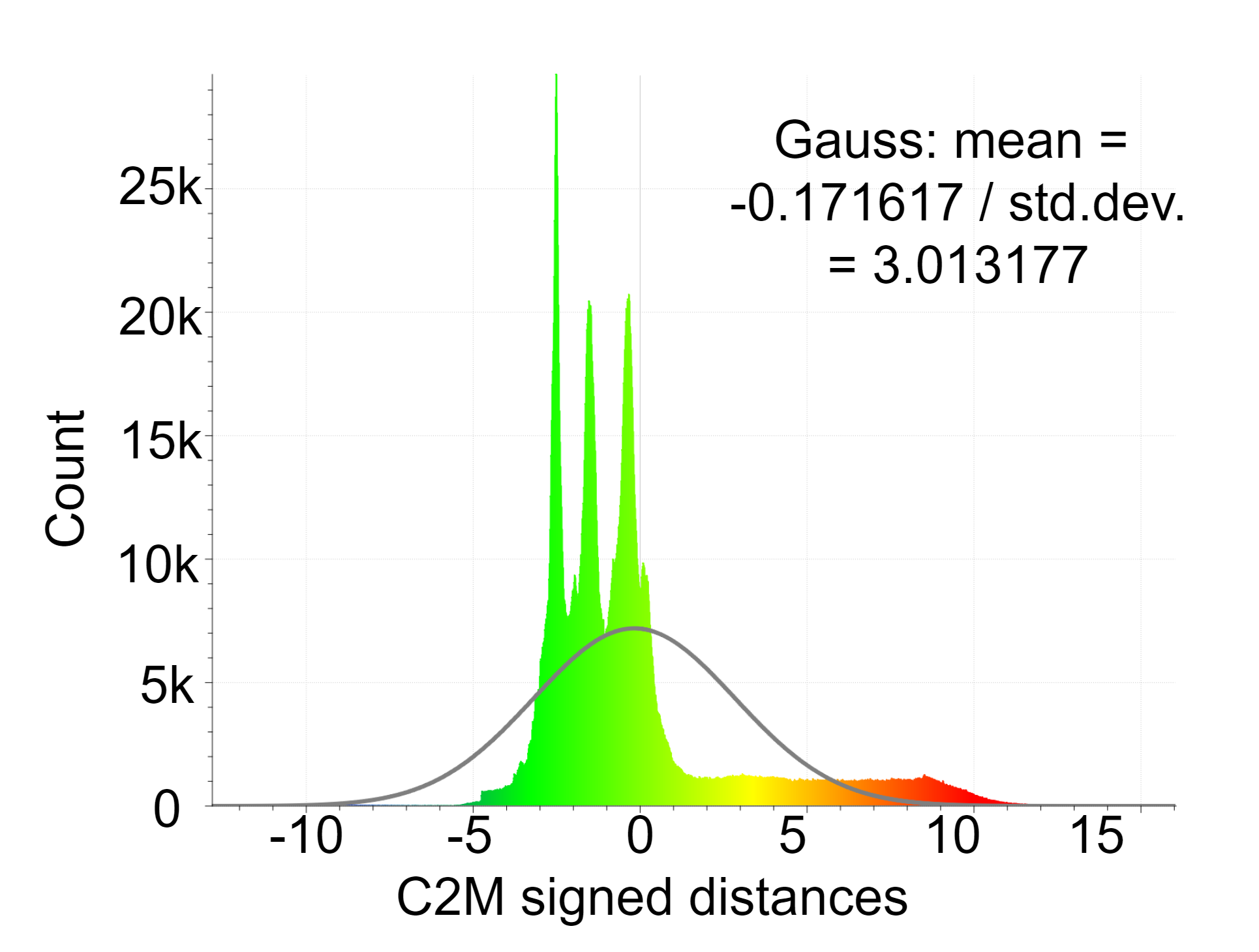}
        \caption{}
        \label{fig:sub2}
    \end{subfigure}
    \hfill
    \begin{subfigure}{0.32\textwidth}
        \centering
        \includegraphics[width=\linewidth]{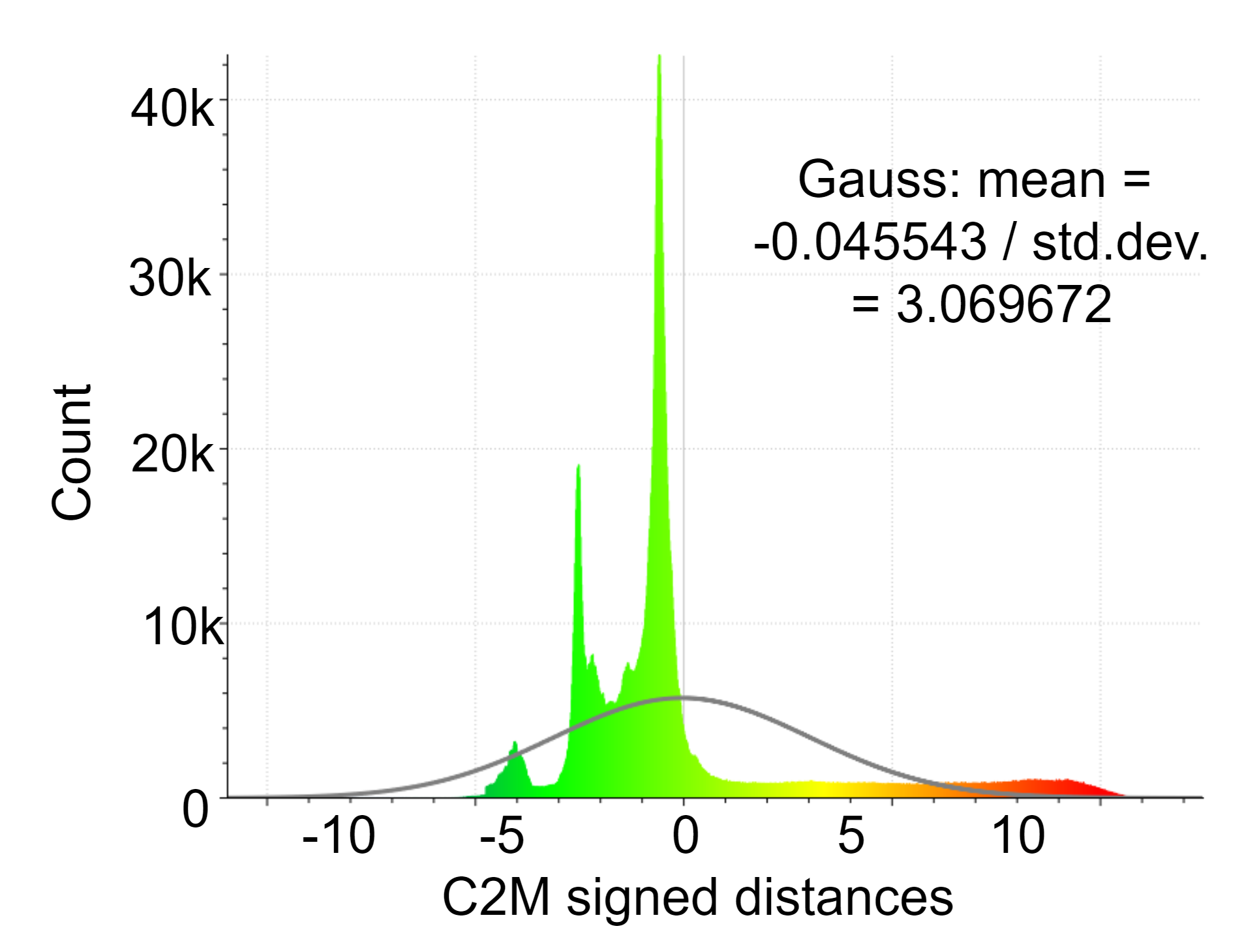}
        \caption{}
        \label{fig:sub3}
    \end{subfigure}
\vspace*{5pt}
    \caption{Distance map obtained by applying Cloud-To-Mesh distance estimation after performing registration using (a) Global ICP, (b) Pose Graph, and (c) Refined Pose Graph.}
    \label{fig:histogram}
\end{figure}

From the images, we observe a concentration of red points below the object, where the holder was located. This discrepancy arises because the holder is not present in the ground truth mesh, resulting in it being perceived as a deviation. The majority of points are green, indicating they are close to zero. To better quantify this results, we plotted a Gaussian distribution of the distances and calculated the mean value, as shown in Figure \ref{fig:histogram}.

As observed, the distance distribution is closer to zero when using the Refined Pose Graph method, with a mean value of $-0.04$. In comparison, the Pose Graph and Global ICP methods performed worse, with mean values of $-0.17$ and $-0.87$, respectively

% \subsection{Comparison of Different Registration Methods}
% % We evaluated several registration methods, including ICP and pose graph. Our metrics for comparison include the accuracy of the registration, speed, robustness to noise and robustness to transformations. 

% \subsection{Comparison of Different Cloud-to cloud Distance Methods}

% \subsection{Comparison with Ground Truth Model}

\section{Conclusions}\label{sec:conclusions}
In this paper, we conducted an evaluation of various multiview registration methods and point cloud distance metrics. We began by generating a synthetic dataset comprising partial scans and complex shape meshes. By applying random rotations and translations to the partial scans and performing registration, we numerically compared different alignment methods based on the degree of rigid transformation applied.

Our findings indicated that the Refined Pose Graph method consistently produced the best results across all ranges of applied rotational and translational transformations. However, achieving these results required careful choice of the algorithm's parameters. Specifically, we identified a relationship between the voxel size, the maximum distance threshold, and the edge prune threshold. Using the Refined Pose Graph method, we developed a novel CloudCompare plugin that incorporates all the functionalities and parameter settings for multi-point cloud fusion. 

Regarding distance metrics, our analysis revealed that as the complexity of the synthetically generated shapes increased, accurately estimating distance became more challenging. This difficulty appeared because all the methods we compared relied on nearest neighbor searches. Nevertheless, the cloud-to-mesh distance metric demonstrated the highest accuracy across various shapes, point cloud densities, and perturbations (noise and holes).

In a real-world scenario, we obtained partial scans using a gantry lab setup for inspection as shown in Figure \ref{fig:setup}, with a static structured light sensor and a moving object. Using the developed CloudCompare plugin, we successfully combined partial scans and performed distance estimation relative to the ground truth.

Future work will focus on enhancing the Refined Pose Graph method by incorporating color information from point clouds. The color information would be beneficial to remove wrong corresponding pairs where the colors of points do not match. Additionally, a systematical analysis is needed to optimally and (semi-)automatically adjust the parameters of the Refined Pose Graph approach.

\bibliographystyle{splncs04}
\bibliography{bibl}

\end{document}